\documentclass[10pt,twocolumn,letterpaper]{article}

\usepackage{iccv}
\usepackage{times}
\usepackage{epsfig}
\usepackage{graphicx}
\usepackage{amsmath}
\usepackage{amssymb}
\usepackage{multirow}
\usepackage{makecell}

\usepackage[pagebackref=true,breaklinks=true,letterpaper=true,colorlinks,bookmarks=false]{hyperref}

\iccvfinalcopy %

\def\iccvPaperID{5327} %

\DeclareMathOperator*{\argmin}{arg\,min}

\ificcvfinal\fi
\begin{document}

\title{Unconstrained Facial Expression Transfer using Style-based Generator}

\author{Chao Yang\qquad Ser-Nam Lim\\Facebook AI}

\maketitle

\begin{abstract}
Facial expression transfer and reenactment has been an important research problem given its applications in face editing, image manipulation, and fabricated videos generation. We present a novel method for image-based facial expression transfer, leveraging the recent style-based GAN shown to be very effective for creating realistic looking images. Given two face images, our method can create plausible results that combine the appearance of one image and the expression of the other. To achieve this, we first propose an optimization procedure based on StyleGAN to infer hierarchical style vector from an image that disentangle different attributes of the face. We further introduce a linear combination scheme that fuses the style vectors of the two given images and generate a new face that combines the expression and appearance of the inputs. Our method can create high-quality synthesis with accurate facial reenactment. Unlike many existing methods, we do not rely on geometry annotations, and can be applied to unconstrained facial images of any identities without the need for retraining, making it feasible to generate large-scale expression-transferred results.
\end{abstract}

\section{Introduction}

\begin{figure}[!h]
\centering
\small
\setlength{\tabcolsep}{1pt}
\begin{tabular}{ccc}
  \includegraphics[width=.15\textwidth]{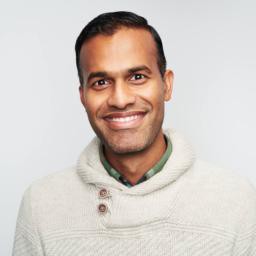}&
  \includegraphics[width=.15\textwidth]{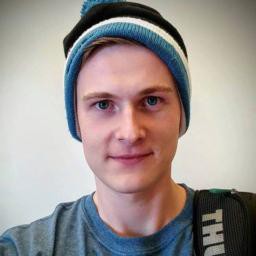}&
  \includegraphics[width=.15\textwidth]{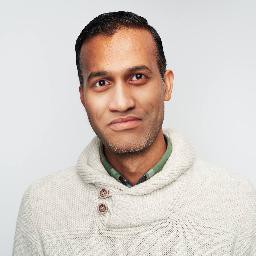}\\
    \includegraphics[width=.15\textwidth]{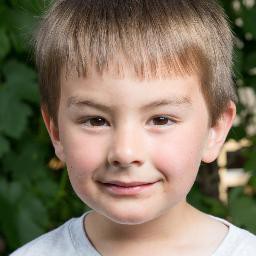}&
  \includegraphics[width=.15\textwidth]{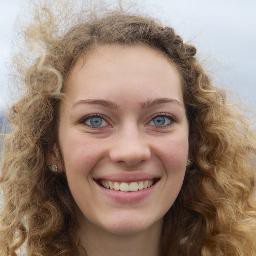}&
  \includegraphics[width=.15\textwidth]{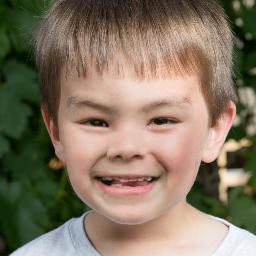}\\
  (a) & (b) & (c) \\
\end{tabular}
\caption{Examples of facial expression transfer between two images. Our method could take any two face images as input and combine the appearance (a) and expression (b) to synthesize realistic-looking reenactment result (c).}.
\label{fig:teaser}
\end{figure}

Many methods have been developed in recent years to render realistic faces or edit specific attributes of face images such as appearances or facial expressions~\cite{kim2018deep,averbuch2017bringing,cao2016real,yeh2016semantic,korshunova2017fast,thies2016face2face,nagano2018pagan,olszewski2017realistic,dale2011video}. Such techniques have found applications in a variety of tasks, including photo editing, visual effects, social VR and AR, as well as the controversial ``DeepFake'' where people create faked images or videos of prominent public figures~\cite{DeepFake}, oftentimes with malicious intent to spread fake news or misinformation. The availability of social media, online image, video portals and public datasets have provided easily-accessible data to facilitate better understanding and modeling of facial attributes at different levels~\cite{karras2017progressive,karras2018style}.

We consider the problem of facial expression transfer in the image space: given two face images, we aim to create a realistic-looking image the combines the appearance of one image and the expression of the other. Existing methods mostly fall into two categories. The first one is purely geometry based, where it fits a blendshape model~\cite{cao2015real} or tracks facial keypoints~\cite{averbuch2017bringing} to warp and synthesize target facial textures. However, they require videos as input to fit the 3D model and track the facial transformations. Also, hidden regions such as teeth and wrinkles are hallucinated by directly transferring from the source expressions and therefore do not account for their different appearances. The second category methods consist of data-driven and learning-based synthesis, leveraging recent advances of deep neural networks. In addition to facing similar issues as geometry-based methods, they are also faced with a scalability issue, as they often require paired training data that are difficult to acquire~\cite{olszewski2017realistic} or need to train separate models for each of the target identity~\cite{korshunova2017fast, DeepFake}. 

To address these issues, we propose a novel expression transfer and reenactment method based on the recent style-based generative adversarial network (GAN)~\cite{karras2018style} (StyleGAN) that was developed to generate realistic-looking human faces. The approach is motivated by several observations. First, as shown in~\cite{zhu2016generative}, deep generative models are capable of learning a low-dimensional manifold of the data. Editing in the latent manifold instead of the pixel space ensures that the image does not fall off the manifold and looks natural and realistic. Second, StyleGAN can learn hierarchical ``style'' vectors that are shown to explain attributes at different levels, from fine attributes such as hair color or eyes open/closed to high-level aspects such as pose, face shape, eyeglasses. At the core of our algorithm is an optimization scheme that infers and combines the style vectors to create a face that fuses the appearance and the expression of two images.

With a pre-trained face model, we could directly apply it to any face and infer its semantic styles in the latent space. Assuming different style layers capture different attributes, we propose an integer linear programming (ILP) framework to optimize the style combinations such that the generated image share the appearance of one face and the expression of another. Unlike previous methods, the inference and optimization scheme can generalize to unseen identities or new faces without re-training the model. Our end-to-end expression transfer system also does not rely on geometry modeling or shape/texture separation, making it more widely applicable to difficult poses or extreme viewpoints. Moreover, our approach is fully automatic and could easily be used to generate results at scale, without compromising on the quality of the results due to the effectiveness of the StyleGAN in modeling natural face distributions.

\begin{figure*}[h]
    \centering
    \includegraphics[width=.98\linewidth]{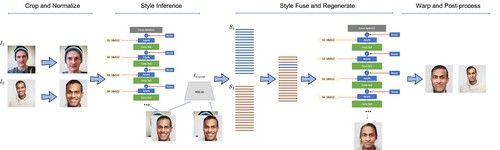}
    \caption{Our system pipeline.} 
\label{fig:pipeline}
\end{figure*}

We test our approach on multiple datasets to show its effectiveness. The results outperform previous geometry-based or learning-based methods both visually and quantitatively. In particular, we demonstrate its application to create large-scale facial expression transfer or ``DeepFake'' data, which can be potentially used to train a more robust detector to safeguard against the misuse of such techniques. 

Our contributions can be summarized as following:
\begin{enumerate}
	\item We introduce an optimization method based on StyleGAN that effectively infer the latent style of a face image. The style is a hierarchical vector that disentangles explanatory factors of the face. 
	\item We propose a unified framework for facial expression transfer based on the latent styles inferred.
	\item We describe a fully automatic, end-to-end system that enables generating high-quality and scalable face reenactment results.
\end{enumerate}

\section{Related Work}

\subsection{Face Reconstruction and Rendering}
Face reconstruction refers to the task of reconstructing 3D face models of shape and appearance from visual data, which is often an essential component in animating or transferring facial expressions. Traditionally, the problem is most commonly tackled using geometry-based method, such as fitting a 3D model to a single image~\cite{blanz2004exchanging,blanz1999morphable} or a video~\cite{cao2014facewarehouse,fyffe2014driving,garrido2014automatic,ichim2015dynamic,shi2014automatic,suwajanakorn2014total,thies2016face2face,wu2016anatomically}. Recently, due to the effectiveness of deep neural networks, learning-based methods have become popular.~\cite{richardson20163d,tewari2017mofa,tuan2017regressing,richardson2017learning,sela2017unrestricted} fit a regressor to predict 3D face shape and appearances from a large corpus of images. ~\cite{sela2017unrestricted} uses an encoder-decoder network to infer a detailed depth image and dense correspondence map which serve as a basis for non-rigid deformation.~\cite{kim2018deep} combines the geometry-based method with learning, which first fits a monocular 3D model and extracts the parameters and then trains a render network using the extracted parameters as input. Unfortunately, its encoder-decoder model is identity-specific, which therefore cannot generalize to new identities. paGAN~\cite{nagano2018pagan} also uses self-supervised training to learn a renderer from mesh deformations and depth to textures. However, when a single image is provided, it needs either the neutral expression or manual initialization. Besides, the result often lacks realism and fine details due to the process of projecting the rendered texture onto a 3D model. 

\subsection{Face Retargeting and Reenactment}
Facial reenactment transfers expressions from a source actor to a target actor. Most of these methods require video as input so as to compute the dense motion fields~\cite{averbuch2017bringing,liu2001expressive,suwajanakorn2015makes} or the warping parameters~\cite{thies2016face2face,thies2016facevr}. Then it uses 2D landmarks or dense 3D models to track the source and target faces. In the case of a single target image, the inner mouth interiors and fine details such as wrinkles are hallucinated by copying from the source, which often leads to uncanny results. Another common approach is to utilize learning-based techniques, especially generative models, for face retargeting.~\cite{olszewski2017realistic} trains a conditional GAN (cGAN) to synthesize realistic inner face textures. However, it requires paired training data, which is difficult to acquire.~\cite{korshunova2017fast} learns face swapping using convolutional neural networks (CNN) trained with content and style loss. The swapping model is also identity-specific and needs to be trained for each target actor. DeepFake~\cite{DeepFake} on the other hand trains an encoder-decoder for each pair of identities to be swapped, where the encoder is shared while the decoder is separate for the source actor and the target actor. Due to the same reason as~\cite{korshunova2017fast}, it also requires laborious training efforts to generalize to new people.

\subsection{Deep Generative Model for Image Synthesis and Disentanglement} 
Deep generative models such as GAN~\cite{goodfellow2014generative} and VAE~\cite{kingma2013auto} have been very successful in modeling natural image distributions and synthesizing realistic-looking figures. Recent advances such as WGAN~\cite{arjovsky2017wasserstein}, BigGAN~\cite{brock2018large}, Progressive GAN~\cite{karras2017progressive} and styleGAN~\cite{karras2018style} have developed better architectures, losses and training schemes that were able to achieve synthesis results of higher resolutions and impressive quality. Due to the nature that generative models learn a low-dimensional manifold from image data, they are often adapted to disentangle latent explanatory factors. UFDN~\cite{liu2018unified} uses conditional VAE (cVAE) to disentangle the domain-invariant and domains specific factors by applying adversarial loss on partial latent encodings. StyleGAN~\cite{karras2018style} modifies GAN architecture to implicitly learn hierarchical latent styles that contribute to the generated faces. However, it remains unclear how to infer these disentangled factors from a given image, and therefore could not be directly applied for manipulating existing data.
\section{Our Approach}

We first define the problem of facial expression transfer. Given an image of the target identity $I_1$, and another image of the source actor $I_2$, we would like to rewrite the facial expression of $I_1$ such that it resembles the expression of $I_2$. Note that unlike~\cite{olszewski2017realistic,DeepFake}, we do not modify the facial appearance of the target identity, instead we only aim to reenact the facial expression of $I_1$ driven by source expression $I_2$. 

\subsection{Overview}
Our pipeline consists of the following components:

\begin{enumerate}
\item Face detect and normalize. We first detect the facial landmarks and use them to crop and normalize the face regions of $I_1$ and $I_2$.
\item Style inference. With the pre-trained StyleGAN model, we iteratively optimize and infer the style vectors of normalized $I_1$ and $I_2$.
\item Style fuse and regenerate. We apply integral linear programming to fuse the two style vectors and regenerate a new image that is combines the appearance of $I_1$ and the expression of $I_2$.
\item Warp and blend. We warp the generated face based on facial landmarks and blend with original $I_1$ rewrite the facial expression so that it agrees with $I_2$.
\end{enumerate}

The pipeline is illustrated in Fig.~\ref{fig:pipeline}.

\subsection{Face Detect and Normalize}

We first detect the facial landmarks using the Dlib detector~\cite{dlib09}, which returns 68 2D keypoints from the image. Given the landmarks, the face could be rectified by computing the rotation based on the eye-to-eye landmark (horizontal) and eye-to-mouth landmark (vertical) directions. We then crop the image based on the eye-to-eye and eye-to-mouth distance, such that the output is a region slightly larger than the face. Finally, we resize the image to 1024x1024, which is the size of the StyleGAN output image. 

\subsection{Style Inference}
\label{sec:style_infer}
\begin{figure}[!h]
\centering
\small
\setlength{\tabcolsep}{1pt}
\begin{tabular}{c}
  \includegraphics[width=.35\textwidth]{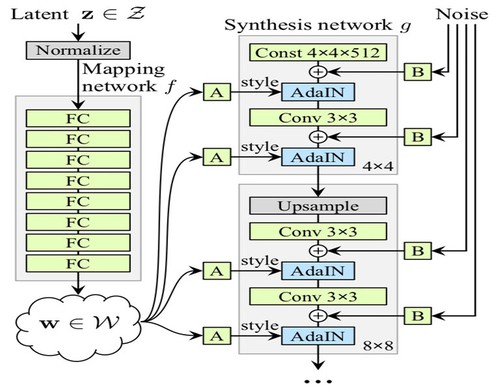} \\
\end{tabular}
\caption{StyleGAN architecture. Figure courtesy of~\cite{karras2018style}.}
\label{fig:styleGAN}
\end{figure}

Given the cropped and rectified image, our goal is to infer its style vector w.r.t. the StyleGAN generator~\cite{karras2018style}. The original StyleGAN (Fig.~\ref{fig:styleGAN}) consists of a mapping network $f$ and a synthesis network $g$. $f$ takes random noise as input and outputs a style vector $s$. $s$ is modeled as 18 layers where each layer is a 512-dimensional vector. The synthesis network takes the style vector and fixed noise as input, where the style vector is used as parameters of adaptive instance normalization ~\cite{huang2017arbitrary} to transform the output before each convolution layer. ~\cite{karras2017progressive} shows that using style vector as layer-wise guidance not only makes synthesizing high-resolution images easier, but also leads to hierarchical disentanglement of local and global attributes.

Our goal is to reverse the process: with a pre-trained model, we aim to find the corresponding style vector $s_I$ of a given image $I$. In this way we could manipulate $s_I$ to change the corresponding attributes of $I$. More formally, our goal is to solve the following objective:

\begin{equation}
s_I = \argmin\limits_{s} D(g(s), I).
\label{eqn:obj}
\end{equation}

Here $g$ is the pre-trained synthesis network with fixed weights, and $D$ is the distance function to measure the similarity between the output image and the original image. Any distance function such as $\ell_1$ or $\ell_2$ could be applied here; however we found using pre-trained VGG network~\cite{simonyan2014very} to compute the perceptual similarity gives best reconstruction results, which is consistent to previous findings that perceptual loss best reflects human sense of similarity~\cite{johnson2016perceptual}. In our experiments, we use the mid-feature layer of VGG-16 pretrained network as the feature extractor before computing the Euclidean distance. More analysis about the distance function and choice of VGG-16 layers are described in ablation study. 

\begin{figure}[!h]
\centering
\small
\setlength{\tabcolsep}{1pt}
\begin{tabular}{c}
  \includegraphics[width=.4\textwidth]{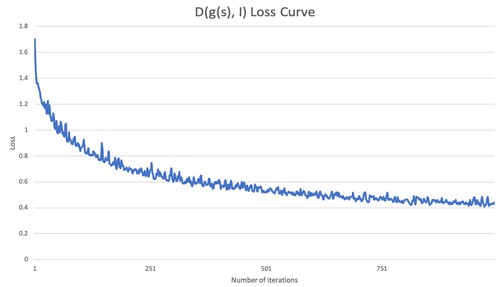} \\
\end{tabular}
\caption{The loss curve as we iteratively optimize~\ref{eqn:obj}.}
\label{fig:loss}
\end{figure}

\begin{figure}[!h]
\centering
\small
\setlength{\tabcolsep}{1pt}
\begin{tabular}{ccccc}
  \includegraphics[width=.09\textwidth]{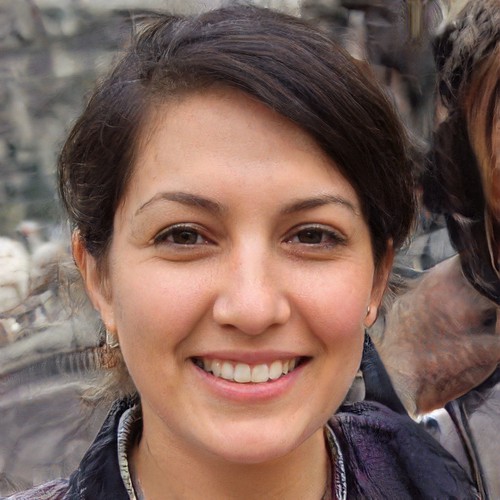} &
  \includegraphics[width=.09\textwidth]{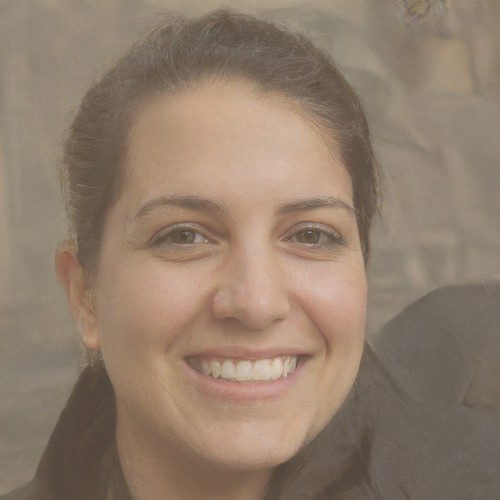} &
  \includegraphics[width=.09\textwidth]{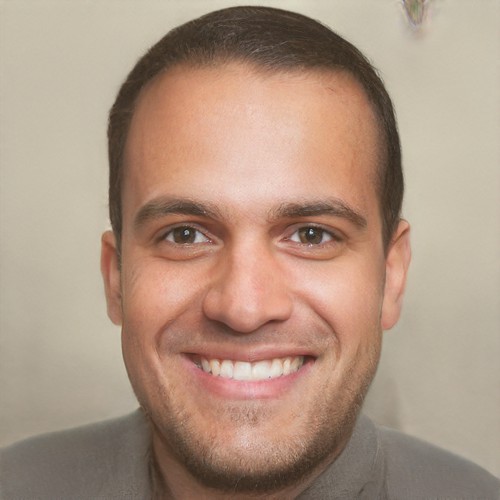} &
  \includegraphics[width=.09\textwidth]{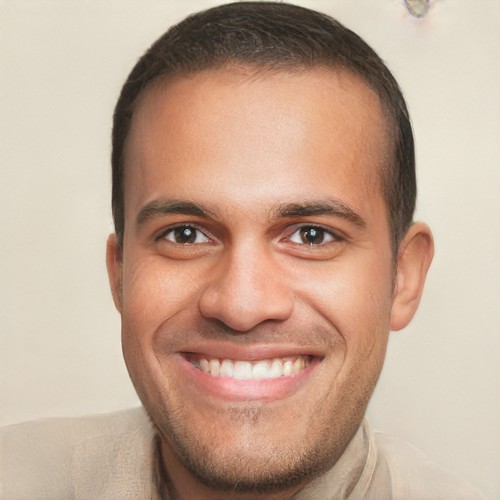} &
  \includegraphics[width=.09\textwidth]{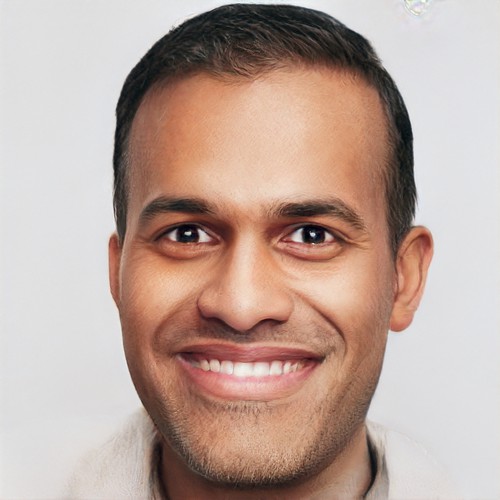} \\
  iter = 0 & iter = 10 & iter = 50 & iter = 100 & iter = 1000 \\
\end{tabular}
\caption{Visualization of $g(s)$ as we iteratively optimize~\ref{eqn:obj}.}
\label{fig:iter}
\end{figure}

Given the image $I$, we iteratively solve for $s_I$ that minimizes Eqn.~\ref{eqn:obj}. $s$ is first initialized as a zero-value style vector and $g(s)$ is a random face. We then compute the error function $D(g(s), I)$, and backpropagate the loss through $g$ to update $s$ using gradient descent. We use learning rate $\mbox{lr}=1$ and for each image, we run 1,000 iterations. Fig.~\ref{fig:loss} shows the an example loss curve of $D$ during iterative optimization, and Fig.~\ref{fig:iter} shows how $g(s)$ evolves to become increasingly similar to $I$. After the optimization is finished, $g(s_I)$ generates a face that is similar to $I$, and $s_I$ can be seen as an approximation of the underlying style vector of $I$. Specific to our task where $I_1$ and $I_2$ are defined as the target identity and source expression image respectively, we solve for their corresponding $s_{I_1}$ and $s_{I_2}$ using the procedures described above.

\subsection{Style Fuse and Regenerate}
\label{sec:style_fuse}
The goal of expression transfer is to come up with a style vector $s_0$ such that $g(s_0)$ shares the appearance of $I_1$ and the expression of $I_2$. More formally, the objective function can be defined as:
\begin{equation}
s_0 = \argmin\limits_{s}D_1(g(s), I_1) + D_2(g(s), I_2).
\label{eqn:obj2}
\end{equation}
Here $D_1$ is the appearance distance and $D_2$ is the expression distance. In order to solve for Eqn.~\ref{eqn:obj2} directly, it requires a well-defined, differentiable expression and appearance distance functions. In~\cite{korshunova2017fast}, $D_2$ is approximated with $\ell_2$ content loss, and $D_1$ approximated with style loss based on Gram matrix. Unfortunately, artifacts in the results show that this does not account for the complexity of facial appearance/expression separation. To ease optimization we constrain the solution space of $s_0$, and assume it lies in the manifold spanned by $s_{I_1}$ and $s_{I_2}$, i.e. $s_0 = \alpha s_{I_1}+\beta s_{I_2}$, where $\alpha$ and $\beta$ are 18x18 diagonal matrix. In other words, each layer of $s_0$ is a linear combination of $s_{I_1}$ and $s_{I_2}$. 

Without defining ad-hoc distance functions, we constrain $\alpha$ to be a 0-1 matrix, and let $\beta=1-\alpha$. This makes Eqn.~\ref{eqn:obj2} an integer linear programming problem and significantly reduce the size of the solution space so that we can heuristically search for the optimal solution. Converting Eqn.~\ref{eqn:obj2} to ILP assumes we always take several layers of $s_{I_2}$ and combine it with the remaining layers of $s_{I_1}$, which is a valid assumption as ~\cite{karras2018style} shows that different style layers correspond to different attributes, and certain layers could represent facial expressions. Our experiments show that using a fixed solution such that $\alpha'=diag(0,0,0,1,1,0,\cdots,0)$ works surprisingly well on a large variety of images. We analyze different combinations of style layers and their effects in ablation study. Finally, we regenerate the image $I_0=g(s_0)$.

\subsection{Warp and Blend}

\begin{figure}[!h]
\centering
\small
\setlength{\tabcolsep}{1pt}
\begin{tabular}{ccccc}
  \includegraphics[width=.09\textwidth]{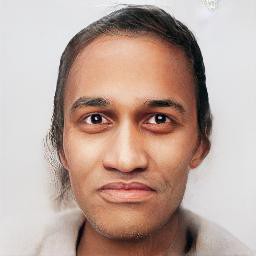} &
  \includegraphics[width=.09\textwidth]{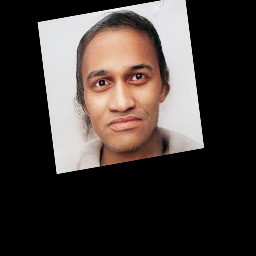} &
  \includegraphics[width=.09\textwidth]{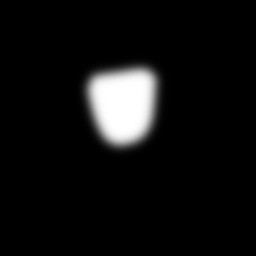} &
  \includegraphics[width=.09\textwidth]{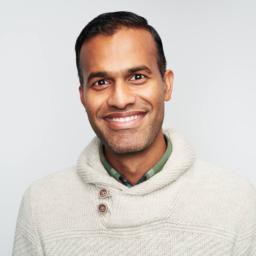} &
  \includegraphics[width=.09\textwidth]{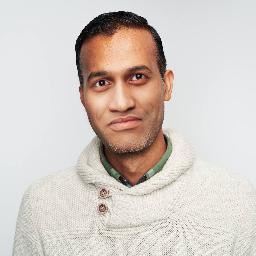} \\
  (a) & (b) & (c) & (d) & (e) \\
\end{tabular}
\caption{Post-processing using warping and blending. (a) generated $I_0=g(s_0)$; (b) warped $I_0$; (c) facial mask; (d) original $I_1$; (e) final composite.}
\label{fig:blend}
\end{figure}

After we regenerate the image $I_0$, we compute a transformation based on the facial landmarks of $I_0$ and original $I_1$ and warp $I_0$ to align with the face region $I_1$. We also compute a mask by taking the convex hull of the facial landmarks and post-processing with Gaussian blur. Finally, we blend warped $I_0$ with $I_1$ using the mask to rewrite the facial expression of $I_1$ (Fig.~\ref{fig:blend}).

\section{Experiments}
\label{sec:setup}
\subsection{Experiment Setup and Results}
We tested our method on several datasets for qualitative illustrations: FFHQ~\cite{karras2018style}, CelebA-HQ~\cite{karras2017progressive}, and random web images. Within each of the dataset, we randomly select pairs of images as the target identity and the source expression. We then infer their style vectors using FFHQ pre-trained StyleGAN model. Note that although the statistics of images vary between datasets, we found that using a uniform pre-trained model works sufficiently well across different data. We then fuse the two style vectors by replacing the expression layers of the target identity with those of the source expression and regenerate the image, which is warped and blended with the target identity as the final output. As discussed in Sec.~\ref{sec:style_fuse}, we always use layer 4 and 5 of a style vector as the representation of expressions. It takes on average 2 minutes to process a pair of images, and most of the time is spent on iterative style vector inference. 

\begin{figure*}[!h]
\centering
\small
\setlength{\tabcolsep}{1pt}
\begin{tabular}{cccc}
  \includegraphics[width=.48\textwidth]{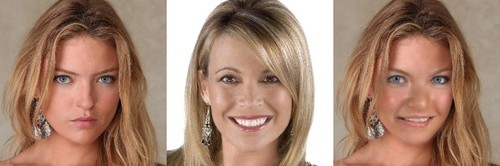} &
  \includegraphics[width=.48\textwidth]{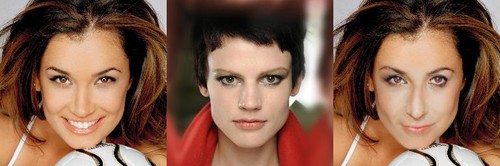} \\
  \includegraphics[width=.48\textwidth]{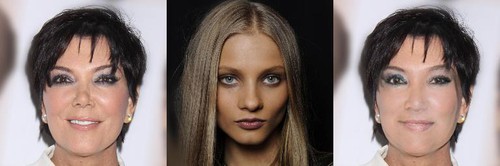} &
  \includegraphics[width=.48\textwidth]{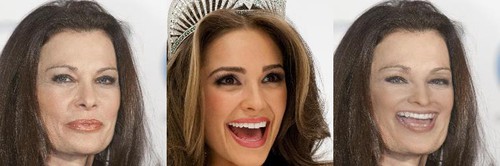}\vspace{4pt} \\ 
  \Xhline{2\arrayrulewidth} \\ 
  \includegraphics[width=.48\textwidth]{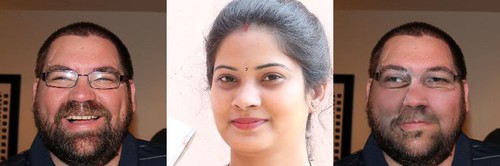} &
  \includegraphics[width=.48\textwidth]{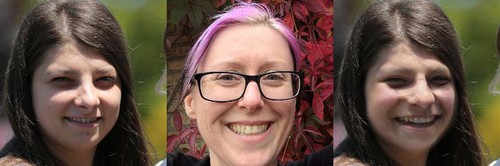} \\
  \includegraphics[width=.48\textwidth]{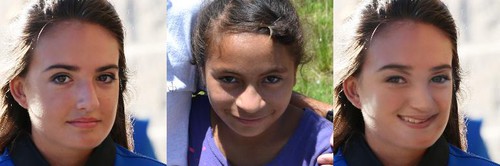} &
  \includegraphics[width=.48\textwidth]{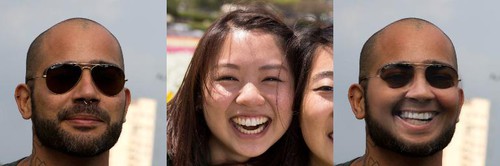} \vspace{4pt} \\
  \Xhline{2\arrayrulewidth} \\ 
  \includegraphics[width=.48\textwidth]{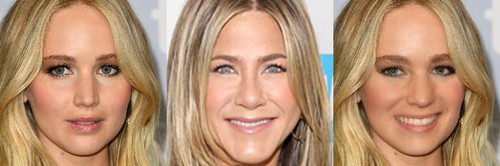} &
  \includegraphics[width=.48\textwidth]{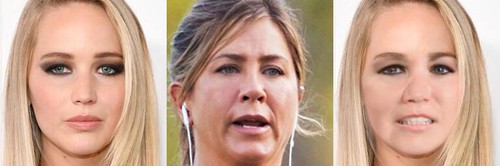} \\
  \includegraphics[width=.48\textwidth]{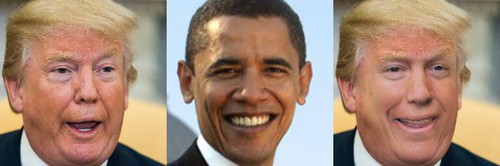} &
  \includegraphics[width=.48\textwidth]{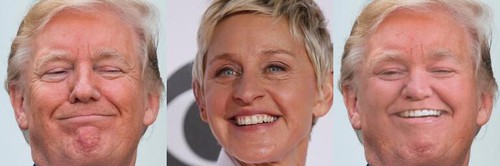} \\
  \includegraphics[width=.48\textwidth]{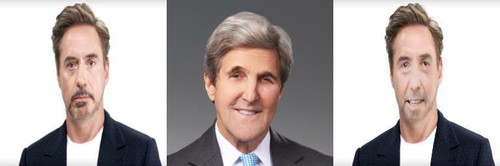} &
  \includegraphics[width=.48\textwidth]{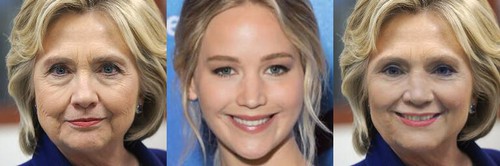} \\
\end{tabular}
\caption{Examples of visual results. Each set consists of identity (left), expression (middle) and result (right). Test images from top to bottom: CelebA-HQ images, FFHQ images, random web images. Note the StyleGAN model is only trained with FFHQ dataset.}
\label{fig:visual}
\end{figure*}

Example visual results are shown in Fig.~\ref{fig:visual}. In most cases, our result looks like plausible human faces and also accurately transfers the expressions of the source actor to the target identity. Other than the mouth regions, it also transfers the eye movements and gaze directions (e.g., Row 3 left). The result keeps the eyeglasses in the target identity but ignore those from the source expression, showing the model is effective in disentangling the appearance/expression attributes. Furthermore, in many cases the appearances of the target identity and the source expression are substantially dissimilar in skin colors, head poses or genders, our model still separates those traits from expressions and achieves satisfactory transferred results. 

\begin{figure}[!h]
\centering
\small
\setlength{\tabcolsep}{1pt}
\begin{tabular}{c@{\hskip .1in}ccccc}
&
\includegraphics[width=.07\textwidth]{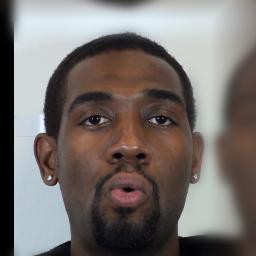}&
\includegraphics[width=.07\textwidth]{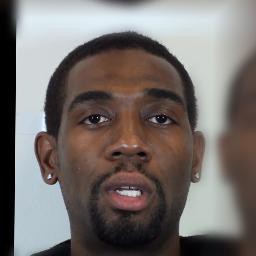}&
\includegraphics[width=.07\textwidth]{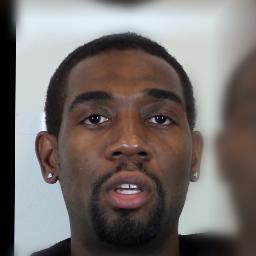}&
\includegraphics[width=.07\textwidth]{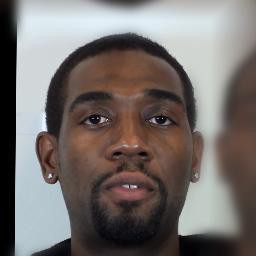}&
\includegraphics[width=.07\textwidth]{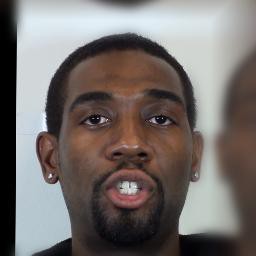}\\ 
\includegraphics[width=.07\textwidth]{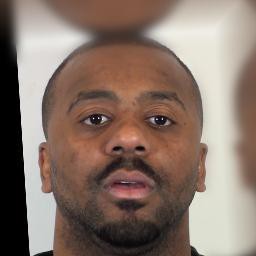}&
\includegraphics[width=.07\textwidth]{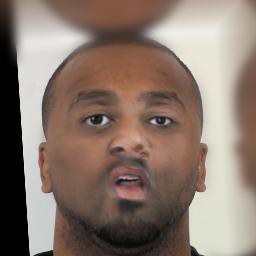}&
\includegraphics[width=.07\textwidth]{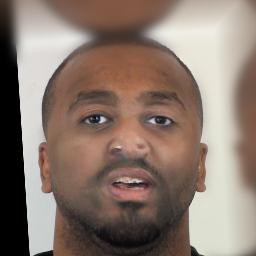}&
\includegraphics[width=.07\textwidth]{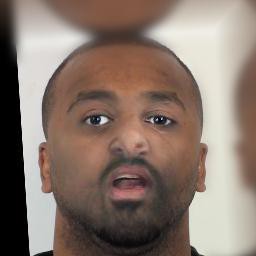}&
\includegraphics[width=.07\textwidth]{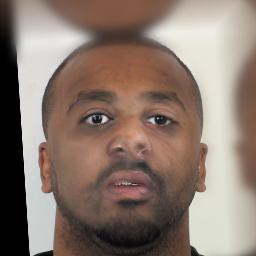}&
\includegraphics[width=.07\textwidth]{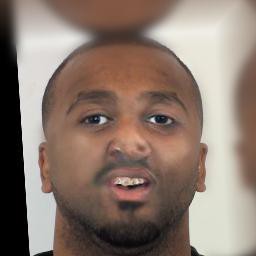}\\
(a)  & 
\end{tabular}
\caption{Expression transfer of videos. (a) target identity; (b) top: source expressions; bottom: transferred results.}
\label{fig:video}
\end{figure}

For quantitative evaluation, we use the videos from~\cite{olszewski2017realistic}. The videos are captured by asking different performers to mimic the expressions and speeches in an instruction video, such that the expressions are frame-wise aligned across different videos. We use one video as source expressions and the first frame of another video as target identity for expression transfer, and then compare the results with the ground truth frames. Qualitative illustrations of Fig.~\ref{fig:video} shows that our model captures and transfers subtle expressions and mouth movements.  Table.~\ref{table:quant} shows quantitative results and compares with~\cite{olszewski2017realistic}. Our results achieve smallest error and highest SSIM, outperforming both cGAN and direct transfer.

\begin{table}[h!]
\begin{center}
  \begin{tabular}{ l | c  c  c}
    \hline
    \textbf{Method} & \textbf{$\ell_1$ Error} &  \textbf{$\ell_2$ Error} & \textbf{SSIM} \\ \hline
    \emph{direct transfer~\cite{olszewski2017realistic}} &  1790 & 211  & 0.815\\ 
    \emph{cGAN~\cite{olszewski2017realistic}} &  1360 & 152  & 0.873\\ 
    \emph{Ours} &  \textbf{1024} & \textbf{136}  & \textbf{0.901} \\ 
    \hline
  \end{tabular}
  \end{center}
  \caption{Numerical comparisons with cGAN and direct transfer. }
  \label{table:quant}
\end{table}

\subsection{Comparisons}
\begin{figure}[!h]
\centering
\small
\setlength{\tabcolsep}{1pt}
\begin{tabular}{cccc}
  \multirow{2}{*}[1cm]{\includegraphics[width=.12\textwidth]{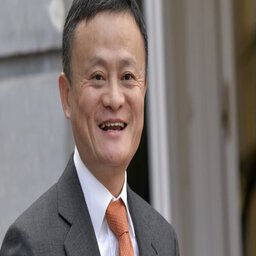}} &
  \includegraphics[width=.12\textwidth]{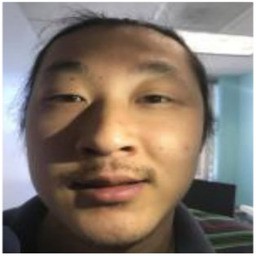} &
  \includegraphics[width=.12\textwidth]{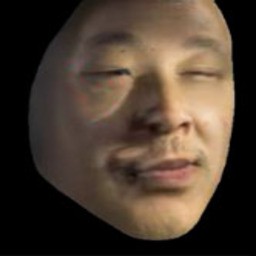} &
  \includegraphics[width=.12\textwidth]{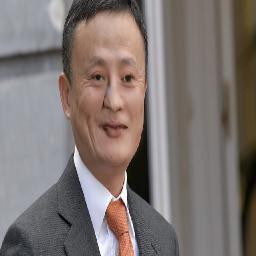} \\
   &
  \includegraphics[width=.12\textwidth]{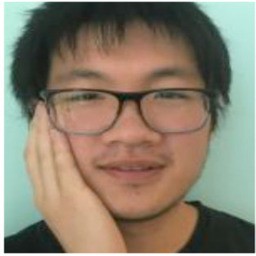} &
  \includegraphics[width=.12\textwidth]{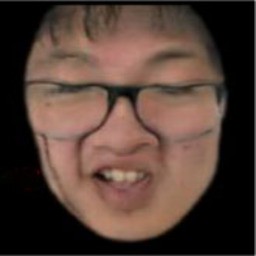} &
  \includegraphics[width=.12\textwidth]{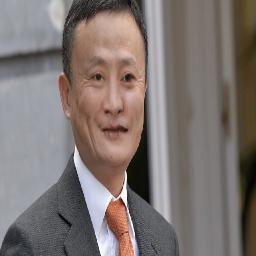} \\
  (a)  & (b) & (c) & (d) \\
\end{tabular}
\caption{paGAN~\cite{nagano2018pagan} comparison. (a) Source identity; (b) Source expressions with shadow (above) and occlusion (below); (c) Texture reconstructions with paGAN; (d) Our expression transfer results. }
\label{fig:pagan}
\end{figure}

\noindent\textbf{Comparison with paGAN~\cite{nagano2018pagan}}
paGAN generates dynamic textures from a single image using a trained network, and the textures could be applied to produce avatars or combine with other faces. Comparing with their method, ours does not need the input to be neutral expressions. Moreover, our model can effectively disentangle expressions from other attributes and is much more robust to handle shadows or occlusions. paGAN on the other hand directly transforms the input textures, so the results largely depend on the input image quality. Fig.~\ref{fig:pagan} shows examples where paGAN fails to reconstruct good textures due to shadow or occlusion while we still manage to capture the expressions and transfer to the target identity accurately. 

\begin{figure}[!h]
\centering
\small
\setlength{\tabcolsep}{1pt}
\begin{tabular}{cccc}
  \includegraphics[width=.12\textwidth]{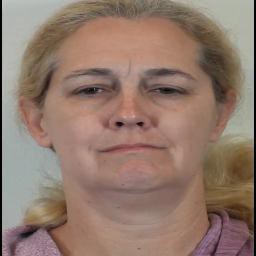} &
  \includegraphics[width=.12\textwidth]{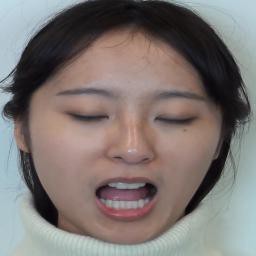} &
  \includegraphics[width=.12\textwidth]{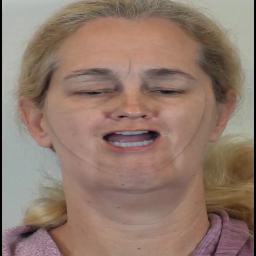} &
  \includegraphics[width=.12\textwidth]{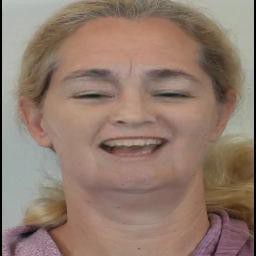} \\
  (a) & (b) & (c) & (d)
\end{tabular}
\caption{Comparison with cGAN. (a) target identity; (b) source expression; (c) their result; (d) our results. Note their mouth interiors are directly copied from source expression.}
\label{fig:cgan}
\end{figure}

\noindent\textbf{Comparison with cGAN~\cite{olszewski2017realistic}} As described in Sec.~\ref{sec:setup}, our results are quantitatively better than~\cite{olszewski2017realistic} when evaluated on videos. One limitation of~\cite{olszewski2017realistic} is that it trains a conditional GAN using registered training pairs, which are difficult to acquire. Another limitation is that it requires the target identity to have neutral expression while our approach does not have any constraint about the facial expression of the target. In addition, for the hidden regions such as mouth interiors, their method directly hallucinates them by copying from the source expressions. This may lead to artifacts as it is from a different person. On the other hand, our approach directly synthesizes the hidden regions, leading to more coherent and realistic results (Fig.~\ref{fig:cgan}). 

\begin{figure}[!h]
\centering
\small
\setlength{\tabcolsep}{1pt}
\begin{tabular}{cccc}
  \includegraphics[width=.12\textwidth]{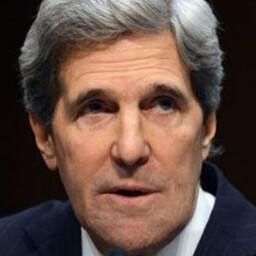} &
  \includegraphics[width=.12\textwidth]{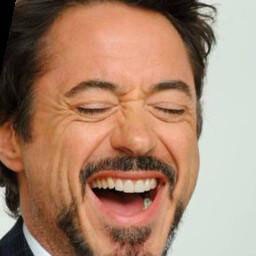} &
  \includegraphics[width=.12\textwidth]{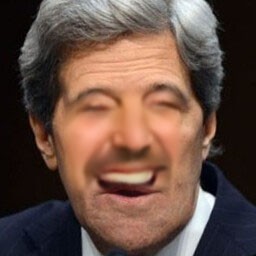} &
  \includegraphics[width=.12\textwidth]{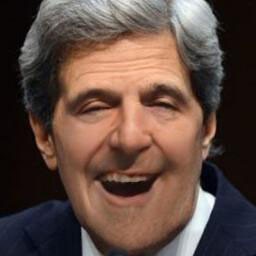} \\
  \includegraphics[width=.12\textwidth]{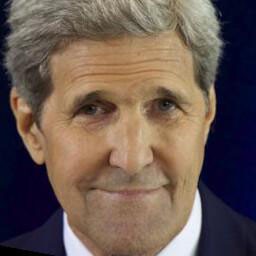} &
  \includegraphics[width=.12\textwidth]{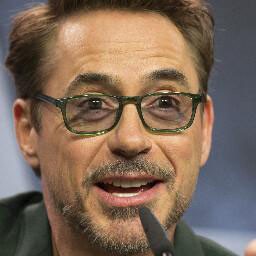} &
  \includegraphics[width=.12\textwidth]{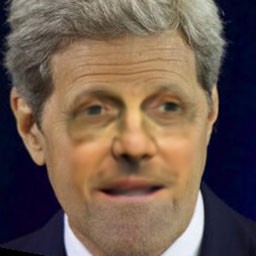} &
  \includegraphics[width=.12\textwidth]{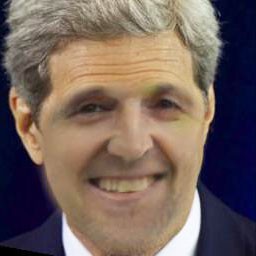} \\
  (a) & (b) & (c) & (d)
\end{tabular}
\caption{Comparison with DeepFake. (a) source identity; (b) source expression; (c) DeepFake result; (d) our result.}
\label{fig:deepfake}
\end{figure}

\noindent\textbf{Comparison with DeepFake~\cite{DeepFake}} DeepFake needs to train a separate model that consists of a shared encoder and different decoders for each pair of identity and expression and therefore is difficult to scale to unseen people. Comparing with their results, our results have higher quality, with more coherent textures and fewer artifacts. We also better disentangle expressions from other attributes such as eyeglasses and mustache (Fig.~\ref{fig:deepfake}). 

\begin{figure}[!h]
\centering
\small
\setlength{\tabcolsep}{1pt}
\begin{tabular}{cccc}
  \includegraphics[width=.12\textwidth]{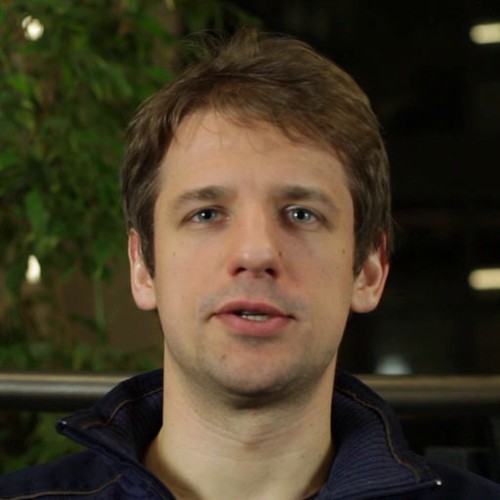} &
  \includegraphics[width=.12\textwidth]{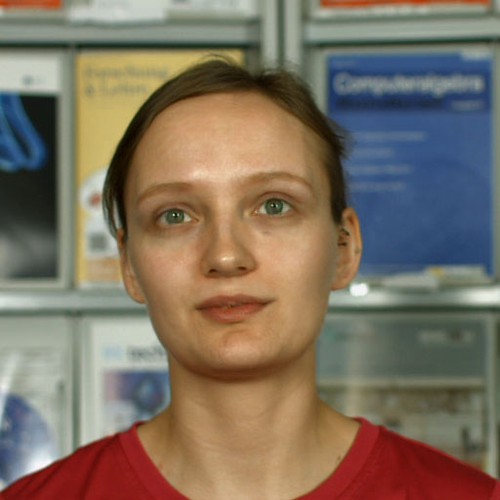} &
  \includegraphics[width=.12\textwidth]{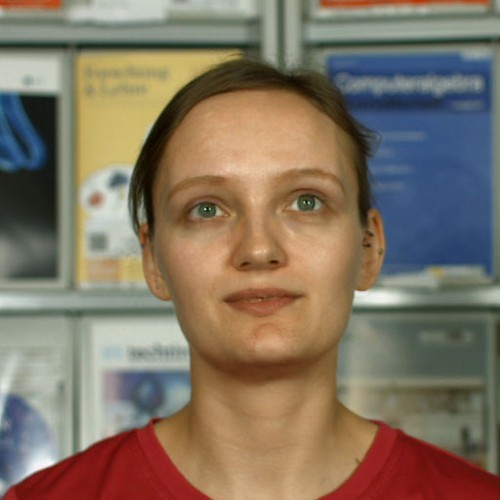} &
  \includegraphics[width=.12\textwidth]{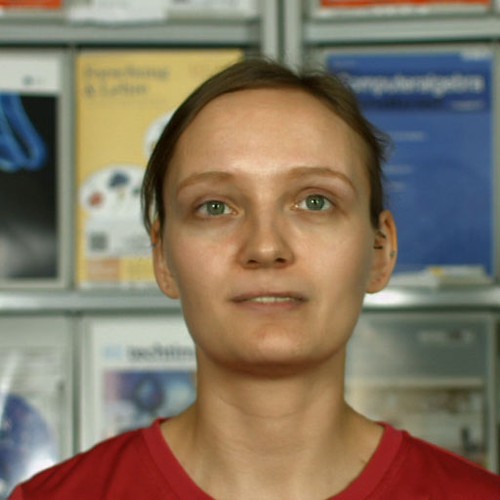} \\
 \includegraphics[width=.12\textwidth]{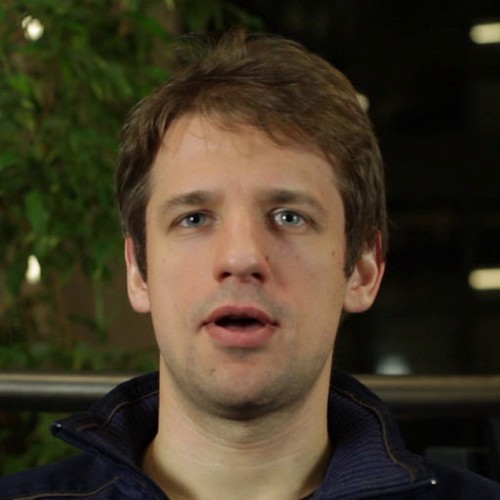} &
  \includegraphics[width=.12\textwidth]{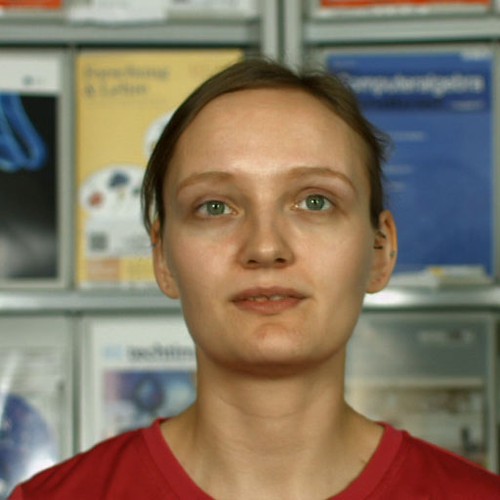} &
  \includegraphics[width=.12\textwidth]{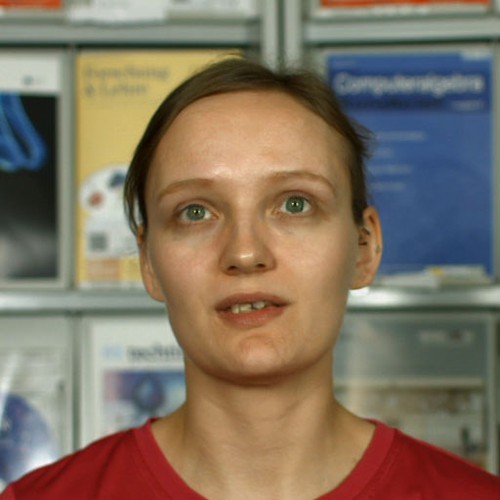} &
  \includegraphics[width=.12\textwidth]{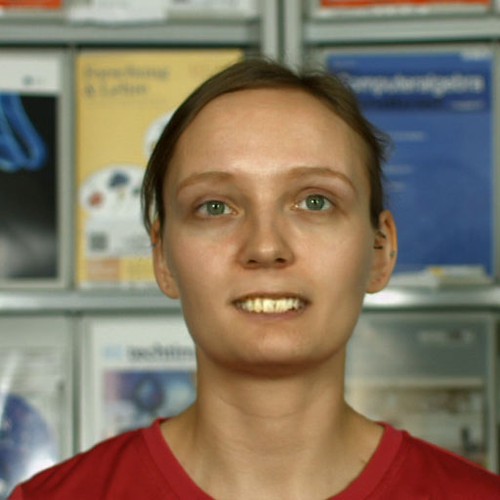} \\
  (a) & (b) & (c) & (d)
\end{tabular}
\caption{Comparison with DVP and Face2Face. (a) source expression; (b) Face2Face; (c) DVP; (d) ours.}
\label{fig:dvp}
\end{figure}

\noindent\textbf{Comparison with Face2Face~\cite{thies2016face2face} and DVP~\cite{kim2018deep}} Face2Face and Deep Video Portraits (DVP) are state-of-the-art facial reenactment approaches. Both methods require video sequence as input and use monocular face reconstruction to parameterize face images. While Face2Face trains multi-linear PCA model for face synthesis, DVP trains a rendering to video translation network to translate modified parameters to images. Similar to~\cite{DeepFake}, both methods need to re-train the model for each new identity, which is cumbersome to scale. On the other hand, our method only requires two images given out model is a generic style extractor that can be applied to any identity. Fig.~\ref{fig:dvp} shows that although all the methods achieve compelling reenactment effects, the facial movements of Face2Face and DVP are modest, while our method can synthesize more drastic expressions. 

\begin{figure}[!h]
\centering
\small
\setlength{\tabcolsep}{1pt}
\begin{tabular}{cccc}
  \includegraphics[width=.12\textwidth]{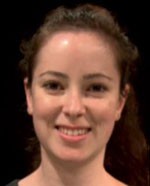} &
  \includegraphics[width=.12\textwidth]{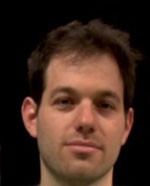} &
  \includegraphics[width=.12\textwidth]{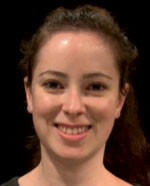} &
  \includegraphics[width=.12\textwidth]{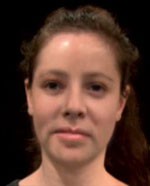} \\
  (a) & (b) & (c) & (d)
\end{tabular}
\caption{Comparison with~\cite{averbuch2017bringing}. (a) target identity; (b) source expression; (c) their result; (d) ours.}
\label{fig:p2l}
\end{figure}

\noindent\textbf{Comparison with ~\cite{averbuch2017bringing}}
We compare our results with~\cite{averbuch2017bringing}, which uses feature tracking and 2D warping to reenact a facial image. Similar to~\cite{olszewski2017realistic}, hallucinating the hidden region such as mouth interior is problematic here as~\cite{averbuch2017bringing} directly copies from the source expression  to the target identity, creating inconsistent appearances. Another issue with~\cite{averbuch2017bringing} is that when the input identity is not neutral expression, the reenactment fails as it is unable to initialize the feature correspondences (Fig.~\ref{fig:p2l}).

\begin{figure}[!h]
\centering
\small
\setlength{\tabcolsep}{1pt}
\begin{tabular}{cccc}
  \includegraphics[width=.12\textwidth]{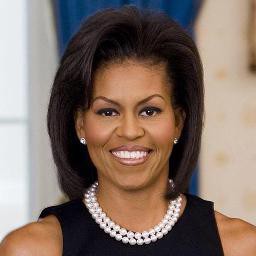} &
  \includegraphics[width=.12\textwidth]{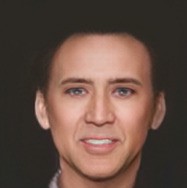} &
  \includegraphics[width=.12\textwidth]{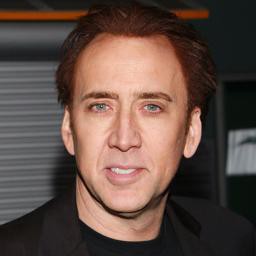} &
  \includegraphics[width=.12\textwidth]{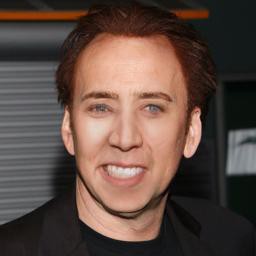} \\
  \includegraphics[width=.12\textwidth]{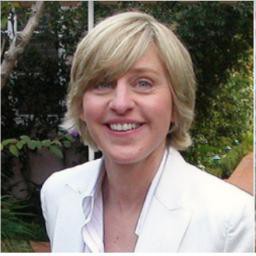} &
  \includegraphics[width=.12\textwidth]{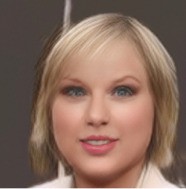} &
  \includegraphics[width=.12\textwidth]{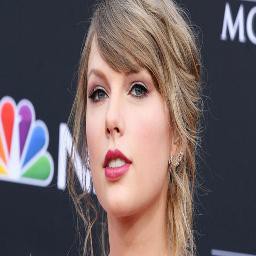} &
  \includegraphics[width=.12\textwidth]{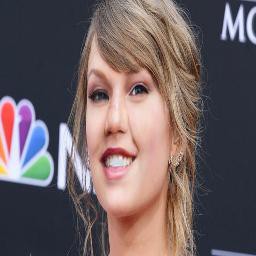} \\
  (a) & (b) & (c) & (d)
\end{tabular}
\caption{Comparison with~\cite{korshunova2017fast}. (a) source expression; (b) ~\cite{korshunova2017fast} face swap result using cage-net or swift-net; (c) random image of Nicolas Cage or Taylor Swift; (d) ours transfer result.}
\label{fig:ffs}
\end{figure}

\noindent\textbf{Comparison with Face Swap~\cite{korshunova2017fast}}
Face Swap~\cite{korshunova2017fast} trains an identity-specific network, such as cage-net or swift-net, to transform any face to appear like the target identity. Naturally, each trained model is limited to swap to a fixed identity. Although we cannot directly compare the results as our goal is not face swapping, we take a random public image from the target identity and apply expression transfer to compare. Fig.~\ref{fig:ffs} shows that our method faithfully transfers the expression to the target identity, while their swapped faces fail to preserve the source expressions.

\subsection{Ablation Study}

\begin{figure}[!h]
\centering
\small
\setlength{\tabcolsep}{1pt}
\begin{tabular}{cccccc}
\includegraphics[width=.08\textwidth]{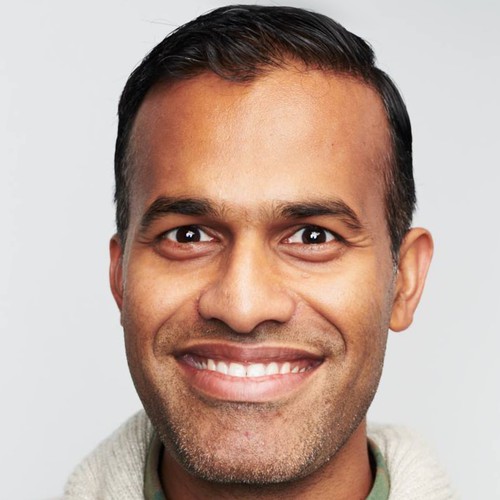} &
  \includegraphics[width=.08\textwidth]{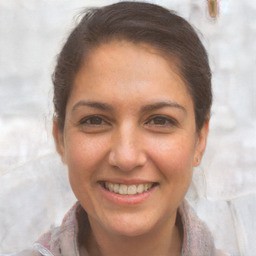} &
 \includegraphics[width=.08\textwidth]{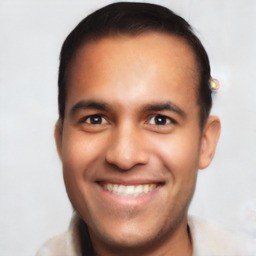} &
 \includegraphics[width=.08\textwidth]{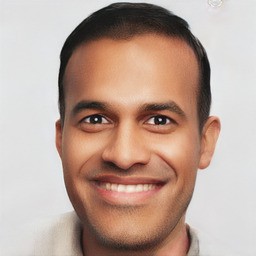} &
 \includegraphics[width=.08\textwidth]{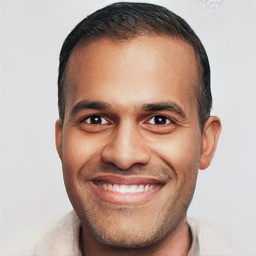} &
 \includegraphics[width=.08\textwidth]{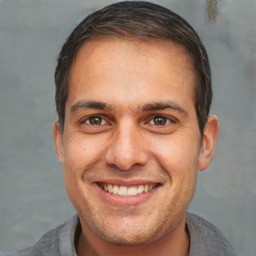} \\
  (a) Input & (b) L=1 & (b) L=3 & (c) L=5 & (d) L=9 & (e) L=13 \\
\end{tabular}
\caption{Reconstruction result when using different layer L of the VGG network as perceptual loss during style vector inference.}
\label{fig:vgg}
\end{figure}

\noindent\textbf{Different reconstruction losses} During iterative style inference, we compute the perceptual loss between the synthesis network output and the original image and back-propagate the loss to update the style vector (Sec.~\ref{sec:style_infer}). We experiment with different layers of the VGG-16 network as the feature extractor when computing the loss function. Fig.~\ref{fig:vgg} shows that using mid-feature layer (L=9) leads to reconstruction of the highest quality and visually most consistent with the original image.  

\begin{figure}[!h]
\centering
\small
\setlength{\tabcolsep}{1pt}
\begin{tabular}{cccc}
  \multicolumn{2}{c}{$I_1$\includegraphics[width=.1\textwidth]{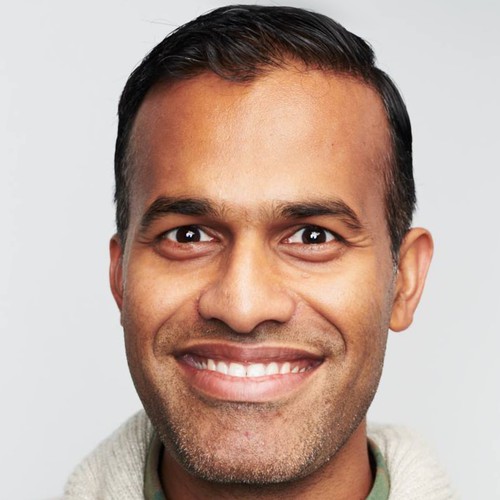}} & \multicolumn{2}{c}{$I_2$\includegraphics[width=.1\textwidth]{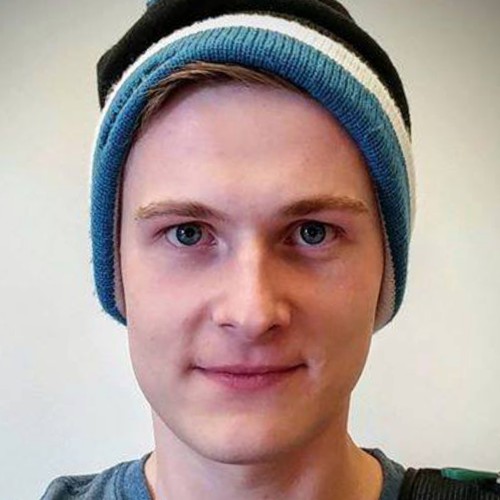}} \\
  $i=2$\includegraphics[width=.1\textwidth]{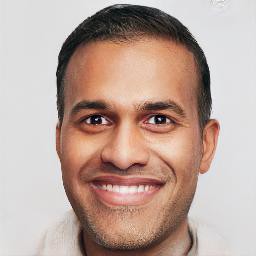} &
  \includegraphics[width=.1\textwidth]{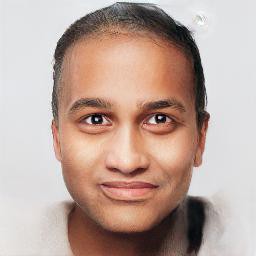} &
  \includegraphics[width=.1\textwidth]{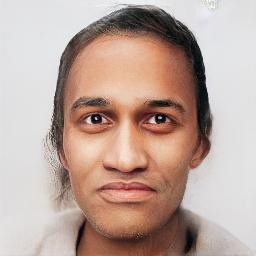} &
  \includegraphics[width=.1\textwidth]{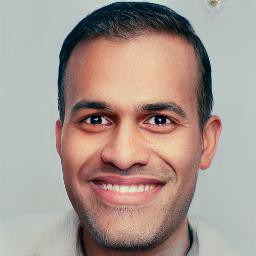} \\ 
  $i=4$\includegraphics[width=.1\textwidth]{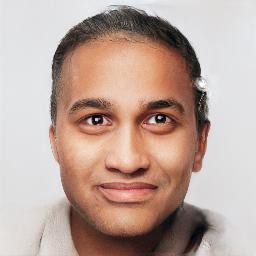} &
  \includegraphics[width=.1\textwidth]{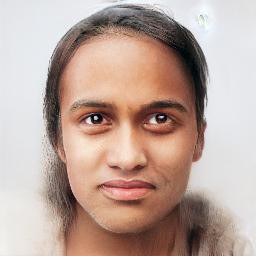} &
  \includegraphics[width=.1\textwidth]{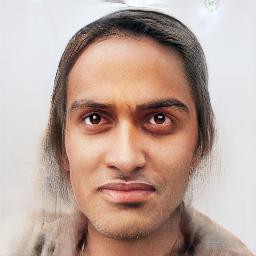} &
  \includegraphics[width=.1\textwidth]{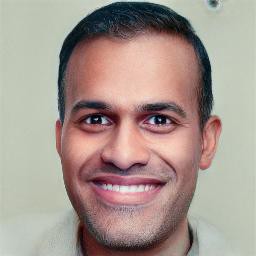} \\ 
  $i=6$\includegraphics[width=.1\textwidth]{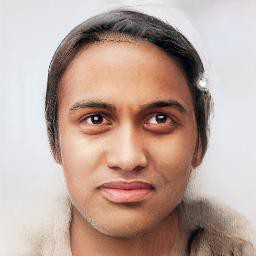} &
  \includegraphics[width=.1\textwidth]{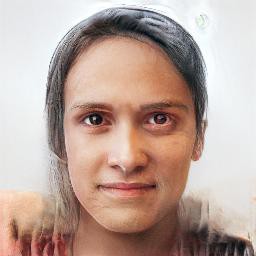} &
  \includegraphics[width=.1\textwidth]{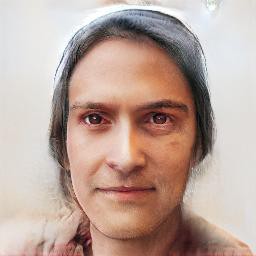} &
  \includegraphics[width=.1\textwidth]{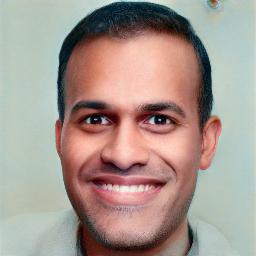} \\ 
  $i=8$\includegraphics[width=.1\textwidth]{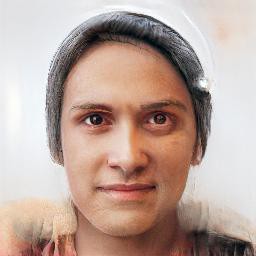} &
  \includegraphics[width=.1\textwidth]{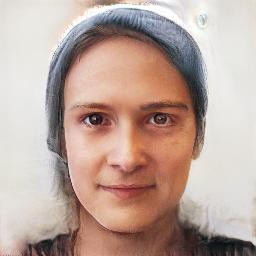} &
  \includegraphics[width=.1\textwidth]{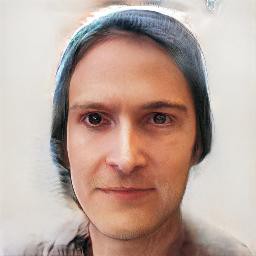} &
  \includegraphics[width=.1\textwidth]{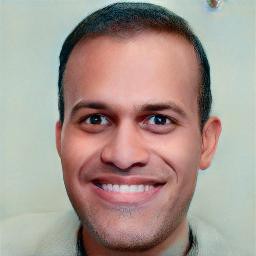} \\ 
   $j=1$ & $j=2$ & $j=4$ & $j=$-1\\
\end{tabular}
\caption{Top row: two images $I_1$ and $I_2$ as input. Bottom rows: replace $i$ layers starting from $j_{th}$ layer of $s_1$ (inferred from $I_1$) with those of $s_2$ (inferred from $I_2$) and regenerate the image. $j=$-1 (last column) indicates replacing the $i$ layers at the end.}
\label{fig:combine}
\end{figure}

\noindent\textbf{Different combinations of style vectors} As described in Sec.~\ref{sec:style_fuse}, we want to search for $s_0$ that is a linear combinations of $s_{I_1}$ and $s_{I_2}$, which are both 18-layer style vectors. Since we constrain the coefficient to be a 0-1 matrix, we heuristically search for different combinations of the two vectors by splicing certain layers of $s_{I_2}$ and replace those of $s_{I_1}$. Fig.~\ref{fig:combine} exhaustively illustrates the regenerated images after we splice the vectors at different locations, and we can see the style vector encodes specific attributes at each layer. For example, the last layer encodes the background style since whenever it changes to $s_{I_1}$ or $s_{I_2}$ the image changes to the corresponding background of $I_1$ or $I_2$. Similarly, the middle layers (layer 8 and 9) can be found to encode the hair color and hat style. Specific our task, it can be observed that layer 4 is associated with expressions: whenever layer 4 is switched to $s_{I_2}$, the regenerated image shows the facial expression of $I_2$. 

\begin{figure}[!h]
\centering
\small
\setlength{\tabcolsep}{1pt}
\begin{tabular}{cc}
  \includegraphics[width=.24\textwidth]{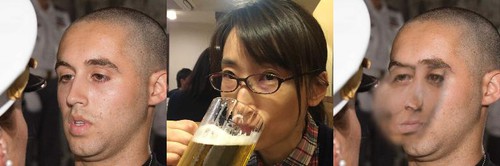} &
  \includegraphics[width=.24\textwidth]{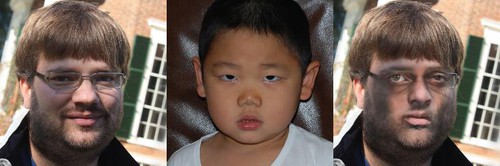} \\
\end{tabular}
\caption{Examples of failure cases. Each set consists of target identity (left), source expression (middle) and result (right).}
\label{fig:failure}
\end{figure}

\noindent\textbf{Failure cases} We also observe some failure cases in our results. For example, when a large portion of the face is occluded, the model fails to recover the complete face and is unable to generate meaningful expressions (Fig.~\ref{fig:failure} left). The output may also look uncanny due to excessive shadows in the source identity (Fig.~\ref{fig:failure} right).

\section{Conclusions}
We propose a simple yet effective expression transfer method based on StyleGAN. Our method can easily apply to any pair of arbitrary face images and transfer the facial expression from one to another. Our approach not only generates compelling results but is also highly scalable and fully automatic. As future work, we are interested in extending our framework to incorporate head pose reenactment to generate more realistic video results. It would also lead to exciting applications if time efficiency could be improved such that expression transfer could run in real-time.

{\small
\bibliographystyle{ieee}
\bibliography{egbib}
}

\end{document}